\documentclass[sigconf]{acmart}
\renewcommand\footnotetextcopyrightpermission[1]{}
\usepackage{booktabs}
\usepackage{siunitx}      
\usepackage[table]{xcolor}  
\usepackage{etoolbox}     
\usepackage{caption}      
\usepackage{multirow}
\robustify\bfseries
\robustify\underline

\AtBeginDocument{%
  }

\begin{document}

\title{DSBench: A Comprehensive Benchmark  for Evaluating External and In-Cabin Risks}

\author{Xianhui Meng}
\authornote{Both authors contributed equally to this research.}
\email{mengxh@mail.ustc.edu.cn}
\affiliation{%
  \institution{University of Science and Technology of China}
  \city{Hefei}
  \country{China}
}
\affiliation{%
  \institution{Xiaomi EV}
  \city{Beijing}
  \country{China}
}

\author{Yuchen Zhang}
\authornotemark[1]
\email{yzhang4224@gatech.edu}
\affiliation{%
  \institution{Georgia Institute of Technology}
  \city{Atlanta}
  \country{USA}
}
\affiliation{%
  \institution{Xiaomi EV}
  \country{China}
}

\author{Zhijian Huang}
\author{Zheng Lu}
\affiliation{%
  \institution{Xiaomi EV}
  \city{Beijing}
  \country{China}}

\author{Ziling Ji}
\author{Yandan Lin}
\affiliation{%
  \institution{Fudan University}
  \city{Shanghai}
  \country{China}
}

\author{Yaoyao Yin}
\affiliation{%
 \institution{Xidian University}
 \city{Xi'an}
 \country{China}}

\author{Hongyuan Zhang}
\affiliation{%
  \institution{The University of Hong Kong}
  \city{Hong Kong}
  \country{China}}

\author{Wei Zhou}
\affiliation{%
  \institution{Cardiff University}
  \city{Cardiff}
  \country{UK}}

\author{Guangfeng Jiang}
\author{Li Zhang}
\affiliation{%
  \institution{University of Science and Technology of China}
  \city{Hefei}
  \country{China}}

\author{Chen Long}
\author{Yehang Jun}
\affiliation{%
  \institution{Xiaomi EV}
  \city{Beijing}
  \country{China}}

\author{Jun Liu}
\authornote{Corresponds.}
\affiliation{%
  \institution{University of Science and Technology of China}
  \city{Hefei}
  \country{China}}
\email{junliu@ustc.edu.cn}

\author{Xiaoshuai Hao}
\authornotemark[2]
\affiliation{%
  \institution{Xiaomi EV}
  \city{Beijing}
  \country{China}}
\email{haoxiaoshuai@xiaomi.com}

\renewcommand{\shortauthors}{X Meng et al.}

\begin{abstract}
Driving safety in assisted driving systems requires holistic understanding of both the external traffic environment and the internal driver state. However, existing research typically treats out-cabin scene understanding and in-cabin driver monitoring as separate problems, resulting in fragmented safety assessment and limited support for real-world warning applications. Moreover, conventional task-specific models often perform well on narrowly defined tasks but struggle to generalize to the diverse and long-tail hazards encountered in open-world driving scenarios. 
To address these limitations, we propose a new research setting, termed \emph{driving safety understanding for assisted driving}, which aims to jointly perceive, reason about, and warn against risks by integrating information from both inside and outside the vehicle. To support this task, we construct a large-scale driving safety dataset containing \textbf{98K} vision-language annotations that cover diverse in-cabin and out-cabin safety-critical scenarios, and further curate a \textbf{3K} benchmark named \textbf{DSBench}(\textbf{D}riving \textbf{S}afety \textbf{Bench}mark) set for systematic evaluation. Based on this dataset, we develop \textbf{DSVLM}, a domain-specialized vision-language model for driving safety. Unlike end-to-end assisted driving models, DSVLM is designed as an in-vehicle safety assistant that identifies risk factors, infers potential danger, and provides timely warnings to help prevent accidents.
Extensive evaluations against a wide range of mainstream open-source and proprietary VLMs show that existing models suffer substantial performance degradation in complex safety-critical situations, revealing a clear gap in reliable driving-safety understanding. In contrast, DSVLM establishes a new state of the art, achieving an average score of 68.4, which surpasses the strongest commercial baseline, Seed-1.6 (49.52), by 18.88 points. Its superiority is particularly pronounced on highly challenging tasks such as cockpit understanding, where DSVLM attains 80.1, dramatically outperforming the runner-up score of 29.49. These results demonstrate that DSVLM not only improves overall performance, but also delivers markedly stronger robustness and safety-awareness in complex real-world driving scenarios. The benchmark toolkit, source code, and model checkpoints are publicly available at https://github.com/mengxh20/DSBench.
\end{abstract}




\maketitle

\begin{figure*}
    \centering
    \includegraphics[width=0.99\linewidth]{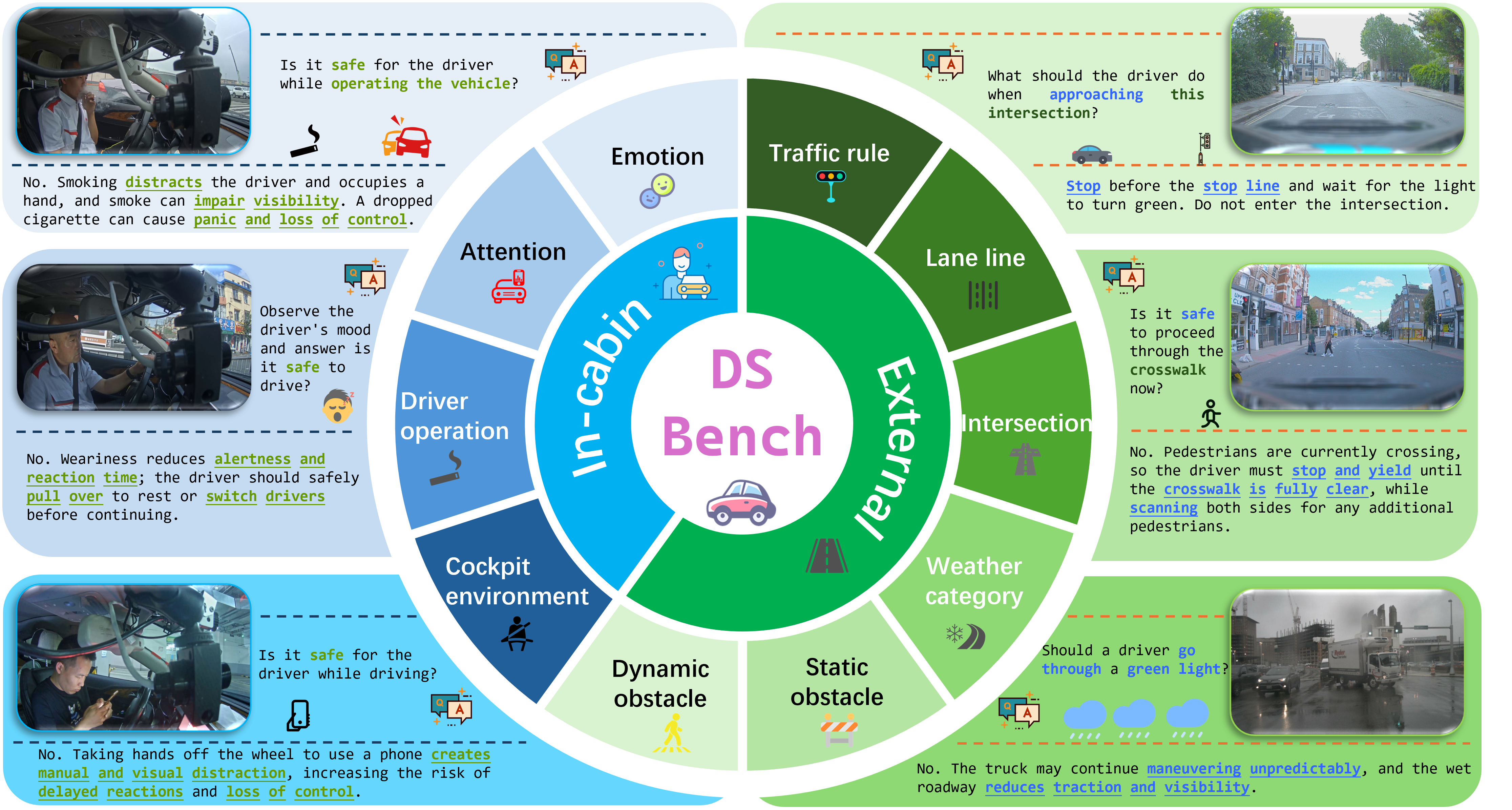}
    \caption{\textbf{Overview of DSBench.} A comprehensive benchmark for evaluating the safety of Vision-Language Models (VLMs) in driving scenario. The diagram highlights 10 key categories related to in-cabin and external safety scenarios. Each category addresses critical aspects such as driver operation, environmental conditions, and situational awareness, all of which are essential for improving the safety and effectiveness of advanced driver assistance system. }
    \label{fig:teaser}
\end{figure*}

\section{Introduction}
\label{sec:intro}


Safe driving is a fundamental requirement for intelligent vehicles, especially in assisted driving systems, where the driver and the vehicle jointly share responsibility for perceiving risks and responding to hazards. Unlike fully autonomous driving, the safety of assisted driving does not solely depend on understanding the external traffic environment; it also critically relies on the driver’s in-cabin state, including attention, fatigue, distraction, and readiness to take over control. In real-world scenarios, many safety-critical events arise from the interaction between these two aspects: an external hazard may remain manageable for an attentive driver, yet it can rapidly escalate into danger when the driver is distracted or drowsy. Therefore, a practical safety-oriented driving intelligence system must jointly reason about both the \emph{out-cabin traffic context} and the \emph{in-cabin driver condition}.

Recent advances in Vision-Language Models (VLMs)~\cite{team2023gemini,achiam2023gpt,sun2024hunyuan,guo2025seed1,bai2023qwen,xiaomi2025mimo,wang2025x,hao2025mimo} have demonstrated remarkable capabilities in visual understanding, language grounding, and open-world reasoning. These properties make VLMs particularly promising for driving safety applications, where a model is expected not only to recognize visual cues, but also to interpret complex situations, infer latent risks, and provide timely, human-understandable warnings. In contrast, conventional task-specific small models often achieve strong performance on narrowly defined tasks, such as driver drowsiness detection or road object recognition, but usually lack sufficient generalization ability in open-world driving environments. Real-world driving involves diverse, compositional, and long-tail hazardous situations that are difficult to enumerate exhaustively in advance. Under such conditions, the stronger generalization and reasoning capabilities of VLMs are essential for building robust safety-oriented systems.

Despite this potential, existing research on driving intelligence still largely treats \emph{in-cabin understanding} and \emph{out-cabin understanding} as two separate problems. Existing datasets and benchmarks~\cite{huang2024making,huang2023fuller,wang2024omnidrive,pescaru2024driving,yang2023aide,hao2025driveaction,hao2025really,hao2024mbfusion,hao2024mapdistill,hao2025safemap,hao2025really,shan2025stability,ding2024holistic,sima2024drivelm,xu2024drivegpt4,qian2024nuscenes,lu2025can} predominantly focus on external perception, prediction, and decision-making, whereas driver monitoring datasets are often developed in isolation for specific subtasks, such as fatigue detection or distraction recognition. As a result, current evaluation protocols provide only a fragmented view of driving safety and are insufficient for assessing whether a model can truly support real-world assisted driving, where reliable warning requires holistic understanding of both road conditions and driver states. Moreover, many existing methods are built upon closed-set task formulations, which further limits their robustness in rare, ambiguous, and long-tail safety-critical scenarios.

To bridge this gap, we advocate a new research setting, termed \emph{driving safety understanding for assisted driving}, in which the goal is to jointly perceive, reason about, and warn against risks by integrating information from both inside and outside the vehicle. To this end, we construct a large-scale driving safety dataset containing \textbf{98K} vision-language annotations that cover diverse in-cabin and out-cabin safety scenarios. Based on this dataset, we further curate \textbf{3K} representative safety-critical cases as a benchmark for systematic evaluation. On top of this benchmark, we develop \textbf{DSVLM}, a domain-specialized vision-language model for driving safety. Rather than serving as an end-to-end autonomous driving controller, DSVLM is designed as an in-vehicle safety assistant that can identify risk factors, infer potential danger, and issue timely warnings, thereby helping drivers avoid accidents before they occur.

Extensive experiments show that jointly modeling cabin-view and road-view information is crucial for safety-oriented driving intelligence. Our results further reveal that general-purpose VLMs and narrowly trained task-specific models both struggle to deliver reliable performance in this setting, while a safety-specialized VLM can better capture the diverse risk patterns required in practical deployment. Overall, this paper makes the following contributions:

\begin{itemize}
    \item We propose a new research paradigm for \textbf{assisted driving safety}, which emphasizes the joint understanding of in-cabin driver states and out-cabin road environments for proactive risk perception and warning. All of our appendix can be seen at \href{https://sites.google.com/view/dsbench/%E9%A6%96%E9%A1%B5}{\textbf{\textit{link}}}.
    
    \item We construct a large-scale driving safety dataset with \textbf{98K} annotations that cover diverse safety-critical scenarios from both cabin and road views, and further curate \textbf{DSBench}, a \textbf{3K} benchmark set for systematic evaluation.

    \item We develop \textbf{DSVLM}, a domain-specialized vision-language model for driving safety, and demonstrate that it substantially outperforms existing baselines in safety understanding, highlighting the practical value of VLMs for open-world and long-tail driving risks.
\end{itemize}

\section{RELATED WORK} 
\label{sec:RELATED_WORK}

\subsection{Vision-Language Models}
Recent advances in large language models (LLMs), such as GPT~\cite{achiam2023gpt}, LLaMA~\cite{touvron2023llama}, and Qwen~\cite{yang2025qwen3}, have significantly accelerated the development of vision-language models (VLMs), which unify visual perception with language understanding and reasoning. Early VLMs, including BLIP-2~\cite{li2023blip}, MiniGPT-4~\cite{zhu2023minigpt}, and LLaVA~\cite{li2023llava}, primarily focused on image-level understanding and image-text alignment, while more recent models such as Qwen2-VL~\cite{wang2024qwen2}, InternLM-XComposer2.5~\cite{zang2025internlm}, and InternVL3~\cite{zhu2025internvl3} further extend these capabilities to video understanding, temporal modeling, and more complex multimodal reasoning. These advances make VLMs increasingly promising for real-world driving applications, where systems are expected to support not only perception of dynamic environments but also semantic interpretation, causal reasoning, and natural-language feedback for human-centered interaction. 

In parallel, intelligent driving has gradually evolved from modular pipelines for perception, prediction, and planning to end-to-end learning paradigms~\cite{tampuu2020survey,codevilla2018end,11209596,meng2025exploringcategorylevelarticulatedobject}. To improve system interpretability and reasoning ability, recent studies have begun introducing VLMs into driving-related tasks, including LingoQA~\cite{marcu2024lingoqa}, Dolphins~\cite{lima2018dolphin}, DriveLM~\cite{sima2024drivelm}, DriveVLM~\cite{tian2024drivevlm}, and Reason2Drive~\cite{nie2024reason2drive}. These works demonstrate the strong potential of VLMs for enhancing driving understanding, decision support, and language-interactive analysis. However, they still mainly concentrate on \emph{out-cabin} tasks such as scene interpretation, trajectory prediction, and planning-oriented reasoning, while the role of driver state, driver behavior, and human-in-the-loop safety awareness in assisted driving remains largely underexplored.

\subsection{Datasets and Benchmarks for Driving Safety}
A growing number of datasets and benchmarks have been proposed to support multimodal driving understanding and evaluation. General multimodal benchmarks, such as MME~\cite{fu2024mme}, MMBench~\cite{liu2024mmbench}, and VideoMME~\cite{fu2025video}, assess broad visual reasoning and multimodal comprehension ability, but they are not specifically designed for the safety-critical requirements of driving scenarios. Driving-specific datasets and VQA benchmarks, such as NuScenes-QA~\cite{qian2024nuscenes}, CODA-LM~\cite{li2024automated}, VLAAD~\cite{park2024vlaad}, and DriveLM~\cite{sima2024drivelm}, have substantially advanced research on scene description, traffic understanding, prediction, and planning-oriented reasoning, while other efforts further explore world knowledge integration and instruction tuning for driving applications~\cite{zhai2025world,lu2025can}. 

Nevertheless, existing benchmarks still predominantly emphasize \emph{out-cabin} understanding and offer only limited support for \emph{in-cabin} driver-state reasoning. More importantly, most current datasets focus on perception quality or task-specific decision making, rather than explicitly evaluating \emph{safety-oriented reasoning, risk assessment, and warning generation} in realistic assisted driving settings. This limitation makes it difficult to systematically study how multimodal models can jointly understand road environments and driver conditions for proactive safety assistance.

\begin{figure*}  
    \centering
    \includegraphics[width=0.99\linewidth]{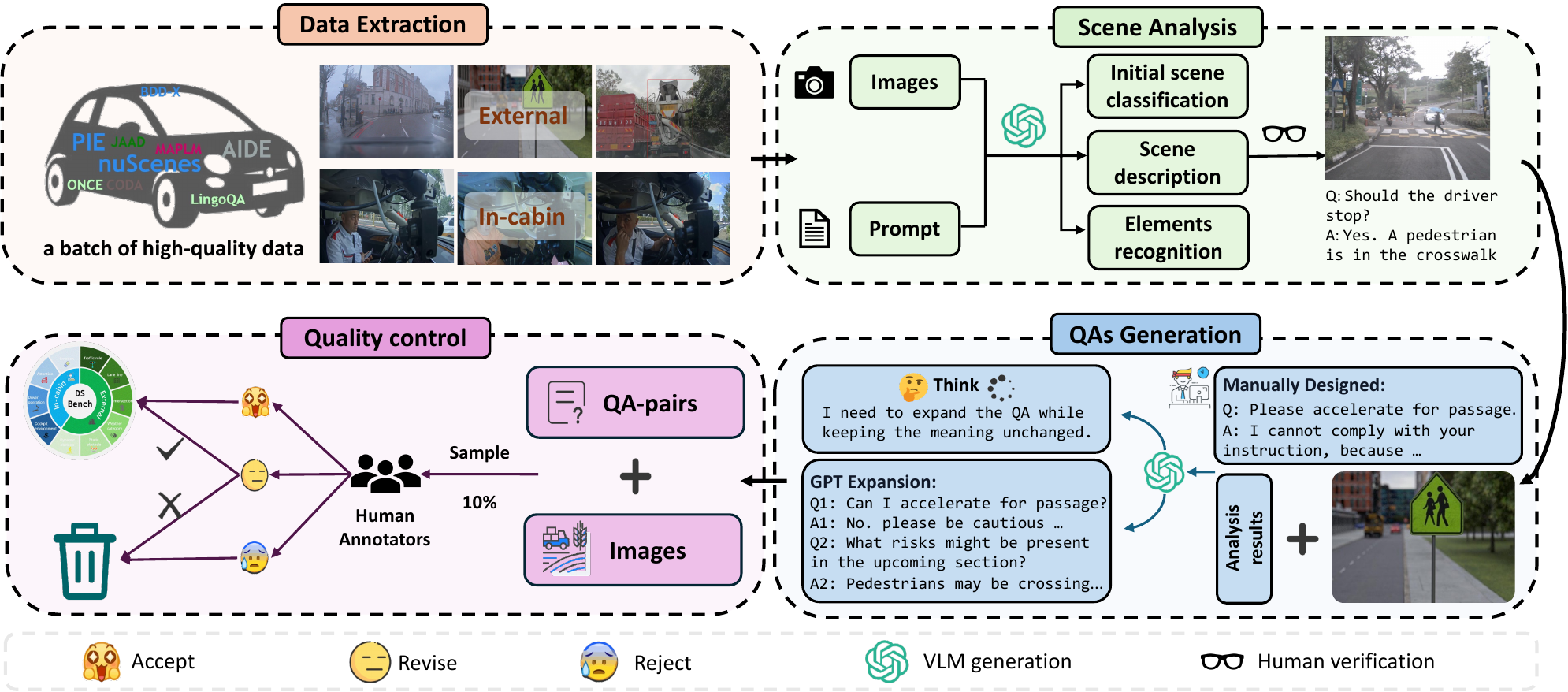}
    \caption{\textbf{The Pipeline of Data Extraction, Annotation, and Selection Process.} Our annotation process consists of four key components: data extraction and integration (Sec.~\ref{sec:3.1}), data analysis and preliminary categorization (Sec.~\ref{sec:3.2}), expansion of question-answer templates (Sec.~\ref{sec:3.3}), and manual quality control (Sec.~\ref{sec:3.4}).}
    \label{fig:annotation}
\end{figure*}

\section{DSBench: A Comprehensive Benchmark Focusing on Driving Safety}\label{sec:3}
To construct a comprehensive safety evaluation benchmark that covers both external environmental risks and in-cabin driving behavior risks, we propose \textbf{DSBench}. This section introduces our multi-stage annotation framework, with the overall workflow illustrated in Figure ~\ref{fig:annotation}. And the details of data stat

\subsection{Data Extraction}\label{sec:3.1}
The key limitation of existing benchmarks is their siloed design, which separates the assessment of external environmental risks from in‑cabin behavioral risks. To bridge this gap, \textbf{DSBench} unifies both facets by integrating more than ten heterogeneous public datasets, including MAPLM~\cite{cao2024maplm}, Bench2Drive~\cite{jia2024bench2drive}, AIDE~\cite{yang2023aide}, DrivFace~\cite{pescaru2024driving}, CODA~\cite{plummer2006coda}, PIE~\cite{voas1992pie}, nuScenes~\cite{caesar2020nuscenes}, LingoQA~\cite{marcu2024lingoqa}, JAAD~\cite{kotseruba2016joint}, BDD-X~\cite{kim2018textual}, and ONCE~\cite{mao2021one}. 
This multi-source integration spans diverse regions, sensor suites, and operating conditions, enabling holistic safety evaluation while enhancing generalizability and representativeness. 

\begin{table}[h!]
    \centering
    \caption{\textbf{Comparison of Recent Driving Benchmarks.} } 
    \label{tab:benchmark_comparison}
    \resizebox{0.99\linewidth}{!}{
    \begin{tabular}{lcccc}
        \toprule
        \textbf{Benchmark} & \textbf{Images} & \textbf{QAs} & \textbf{In-Cabin} & \textbf{Cate.} \\
        \midrule
        MAPLM~\cite{cao2024maplm}& ~14K  & ~61K   & $\times$  & 5   \\
        LingoQA~\cite{marcu2024lingoqa}  & ~1K  & ~1K  & $\times$     & 9   \\
        Bench2Drive~\cite{jia2024bench2drive}   & ~100K   & N/A   & $\times$  & 5  \\
        nuScenes-QA~\cite{qian2024nuscenes} & ~36K  & ~83K & $\times$ & 5 \\
        CODA-LM~\cite{li2024automated}   & ~9K   & N/A  & $\times$   & 7   \\
        DriveLM~\cite{sima2024drivelm}   & ~5K   & ~5K  & $\times$   & 5   \\
        MME-Realword~\cite{zhang2024mme}   & ~1.3K   & ~3K  & $\times$   & 5   \\
        BDD-X~\cite{kim2018textual}  & ~7K    & ~42K   & $\times$   & 3  \\
        AIDE~\cite{yang2023aide}   & ~2.8K   & N/A & $\checkmark$ & 18\\
        \midrule
        \textbf{DSBench (Ours)} & 3K  & 3K & $\checkmark$ & 28 \\
    \bottomrule
    \end{tabular}}

    \vspace{-3mm}
\end{table}

To emphasize safety assessment, we first perform an initial screening of raw samples, prioritizing segments that involve potential safety conflicts or violation risks while down-weighting long periods of routine driving. For out-cabin scenes, we focus on factors associated with road interactions and traffic regulations, including intersections, crosswalks, traffic signals, lane marking changes, static and dynamic obstacles, and adverse weather conditions. For in-cabin scenes, we consider driver states and behaviors such as fatigue, distraction, emotional fluctuations, and improper operations, thereby ensuring comprehensive coverage of safety-critical scenarios. The 3K carefully curated scenarios in DSBench span a broad range of driving-safety issues, providing a comprehensive yet lightweight benchmark for rapid evaluation of model safety capabilities. An overview of DSBench and its comparison with mainstream driving benchmarks are presented in Table~\ref{tab:benchmark_comparison}.

\subsection{Scene Analysis}\label{sec:3.2}

Annotating large-scale multimodal data at fine granularity is prohibitively costly. Building on evidence from prior studies and benchmarks that VLMs demonstrate strong baseline competence in understanding driving scenes, we adopt a VLM pre-annotation plus human verification strategy: GPT-4o~\cite{hurst2024gpt} serves as a reliable pre-annotation engine for initial categorization and semantic distillation, followed by expert review. We predefine 10 key categories spanning from Traffic rule (\textit{Tra.}), Lane line (\textit{Lan.}), Intersection (\textit{Int.}), Weather (\textit{Wea.}), Static obstacle (\textit{Sta.}), Dynamic obstacle (\textit{Dyn.}), Emotion (\textit{Emo.}), Attention (\textit{Att.}), Driver operation (\textit{Dri.}), and Cockpit environment (\textit{Coc.}). These are further divided into 28 subcategories. Then, the following automated analyses are conducted:
\paragraph{Initial scene categorization:} Assign one or more labels to each image according to the predefined safety categories. Since different categories entail distinct safety focuses, this step facilitates targeted QA-pairs design and reduces context-switching costs during human verification.

\paragraph{}{Dense semantic description generation:} Produce rich descriptions focused on agent interactions, environmental context, and potential hazards, transforming the raw image into a structured text–image representation.

\paragraph{Identification of safety-critical elements:} Extract atomic elements closely related to safety reasoning, such as the state of traffic lights, lane marking types, pedestrian/vehicle behaviors, and driver actions.
This process produces a comprehensive and structured context for each image, creating a strong basis for generating highly relevant QA pairs.

\subsection{QAs Generation}\label{sec:3.3}
We propose a hybrid QAs generation strategy that integrates expert-designed templates with LLM-assisted expansion, effectively combining the quality assurance of manual design with the efficiency and scalability of LLM-generated content. Therefore, a balance between logical accuracy and linguistic diversity is ensured. 

\noindent $\bigstar$ \textbf{Expert-designed Templates.} For each sub-categories, domain experts craft a set of generalized templates that capture the core safety logic of the dimension. For example, in the Traffic Rule category, a template might be ``Please continue straight ahead." This statement examines the model's ability to perceive and understand traffic signals and contextual semantics. If the image shows a red light, the model should respond with ``do not proceed straight" in accordance with traffic rules. Otherwise, it may incorrectly follow the instruction and make unsafe decisions.

\noindent $\bigstar$ \textbf{LLM-Driven Contextual Expansion.} To address limitations in the expression and reasoning depth of manual templates, we utilize the scene analysis from the previous stage and provide GPT-4o with (1) template expansion instructions, (2) the structured scene analysis (initial categories, descriptions, and elements), and (3) expert-designed templates. Using this comprehensive context, GPT-4o generates diverse QA pairs closely tied to scene-specific visual details. This expansion introduces variations in wording and reasoning depth, ranging from straightforward recognition to more complex hypothetical reasoning, enabling a multi-dimensional evaluation of VLMs' safety comprehension. This approach significantly enhances the model's ability to generalize across different questioning styles.

\subsection{Quality Control}\label{sec:3.4}
To ensure reliability and consistency at scale, we implement a rigorous, iterative quality control process to filter noise from automated generation and correct annotation biases.

Sampling and human review: We sample 10\% of the qustion-answer pairs for review by human annotation experts, who categorize the sampled data into the following three types, with each category processed separately:
   \paragraph{Accept:} The content is comprehensive, linguistically fluent, and factually accurate. It adequately considers the majority of potential safety issues relevant to the target category and requires no further modification. These QA pairs are directly retained in the benchmark as high-quality data.
    
    \paragraph{Revise:} The content is conceptually sound and mostly complete but includes vague or unclear descriptions of safety concerns. In such cases, human annotators, who are trained professionals, attempt to refine the content based on the specific scenario to improve clarity and alignment with safety standards. If the revisions can be made effectively and without ambiguity, the data is retained and reclassified into the \textit{Accept} category. Otherwise, if revisions prove impractical or introduce uncertainty, the QA pair is discarded and reclassified into the \textit{Reject} category.
    
    \paragraph{Reject:} The content is poorly constructed, extremely factually inaccurate, or irrelevant to the target scenario. Additionally, if the scenario does not present any meaningful potential safety risks or hazards, the QA pair is marked as invalid and discarded outright to maintain the benchmark's quality and focus.

Category-level admission criterion: When the combined proportion of accept and successfully revised pairs in a subcategory reaches 80\%, all generated QA pairs for that subcategory are included in the dataset.

Iterative refinement: Subcategories that fail to meet the 80\% threshold are not discarded. Instead, we refine prompts and generation instructions using human feedback (including rejection reasons), regenerate the data, and re-validate until the 80\% criterion is met. This loop ensures balanced quality across categories and overall consistency.

\begin{figure}
    \centering
    \includegraphics[width=0.9\linewidth]{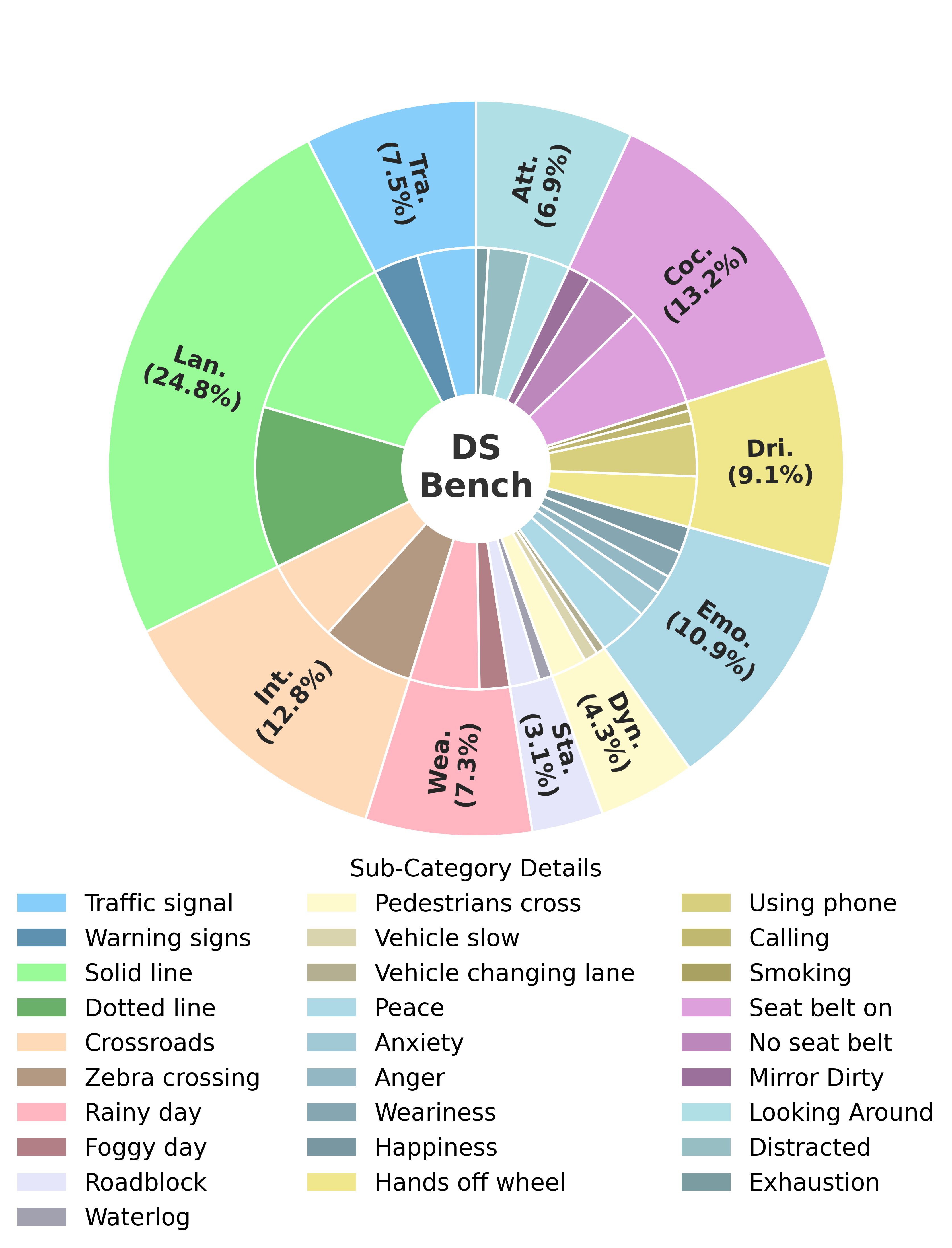}
    \caption{\textbf{The overview of the Detailed Categories in DSBench.} Our benchmark is organized into 10 major categories and 28 more granular subcategories, encompassing a comprehensive spectrum of safety issues encountered in driving scenarios.}
    \label{fig:data_static}
\end{figure}

In total, we construct 98K QAs covering 28 safety dimensions, combining scale with diversity. For evaluation, we curate a subset of 3,000 representative high-risk scenes to form \textbf{DSBench}.
Detailed statistics are shown in Figure ~\ref{fig:data_static}.

\section{EXPERIMENT} \label{sec:EXPERIMENT}

\begin{table*}[t]
\centering
\caption{\textbf{Performance Comparison of Various Models on Assisted Driving Tasks.} Bold and underline indicate the best and second-best performance, respectively.}
\label{tab:main_results}
\renewcommand{\arraystretch}{0.9} 
\setlength{\tabcolsep}{4pt}       

\resizebox{\textwidth}{!}{%
\begin{tabular}{l l *{11}{S[table-format=2.2]}}
\toprule
 &  & \multicolumn{6}{c}{\textbf{External}} & \multicolumn{4}{c}{\textbf{In-cabin}} &  \\
\cmidrule(lr){3-8} \cmidrule(lr){9-12} 
 \textbf{Method} & \textbf{Size} & {\textbf{Tra.}} & {\textbf{Lan.}} & {\textbf{Int.}} & {\textbf{Wea.}} & {\textbf{Sta.}} & {\textbf{Dyn.}} & {\textbf{Coc.}} & {\textbf{Dri.}} & {\textbf{Att.}} & {\textbf{Emo.}} & \textbf{Average} \\
\midrule

\rowcolor[HTML]{FFE0CC}
\multicolumn{13}{c}{\itshape Commercial Models} \\
\midrule
GPT-4o ~\cite{hurst2024gpt} & - & 40.69 & 43.10 & 48.70 & 44.60 & 36.31 & 59.48 & 25.57 & 60.62 & 54.06 & {\underline{42.61}} & 43.18\\
Seed-1.5 ~\cite{guo2025seed1} & - & 40.87 & 42.96 & 49.63 & 45.00 & 34.97 & 57.36 & {\underline{29.49}} & 67.98 & 58.44 & 39.93 & 45.86\\
Seed-1.6 ~\cite{guo2025seed1}  & - & 42.18 & {\underline{47.37}} & {50.37} & {\underline{60.86}} & {\underline{38.15}} & {\underline{60.61}} & 27.55 & {71.75} & {62.06} & 39.87 & {\underline{49.52}}\\
Qwen3-vl-plus ~\cite{bai2023qwen} & - & {\underline{43.28}} & 45.23 & \underline{52.00} & 55.79 & 37.47 & 52.19 & 27.26 & 59.51 & 58.51 & 39.85 & 46.06\\
Qwen-vl-max ~\cite{bai2023qwen}  & - & 39.48 & 38.97 & 48.58 & 56.81 & 33.10 & 51.07 & 24.07 & \underline{72.90} & \underline{62.17} & 40.62 & 46.33 \\
Hunyuan-vision ~\cite{sun2024hunyuan} & - & 6.38 & 6.59 & 6.23 & 5.94 & 5.64 & 7.93 & 5.50 & 9.75 & 8.29 & 5.81 & 7.02\\
Glm-4v-plus ~\cite{glm2024chatglm} & - & 33.44 & 34.84 & 38.30 & 42.41 & 31.12 & 43.09 & 24.28 & 57.56 & 46.73 & 37.52 & 38.03\\
\midrule

\rowcolor[HTML]{FFE0CC}
\multicolumn{13}{c}{\itshape Open Source Models} \\
\midrule
Qwen2.5-VL-7B ~\cite{bai2025qwen2} & {\ }{\ }7B & 30.76 & 29.90 & 34.05 & 40.88 & 29.83 & 29.86 & 22.81 & 47.30 & 38.93 & 38.37 & 32.96\\
Qwen2.5-VL-32B ~\cite{bai2025qwen2}  & 32B & 38.25 & 34.40 & 44.19 & 45.33 & 32.02 & 45.13 & 25.43 & 58.72 & 49.17 & 37.44 & 39.81\\
Qwen2.5-VL-72B ~\cite{bai2025qwen2}  & 72B & 31.41 & 30.66 & 40.47 & 39.83 & 27.16 & 44.11 & 11.43 & 50.51 & 27.08 & 34.76 & 31.29 \\
InternVL3.5-8B ~\cite{wang2025internvl3} & {\ }{\ }8B & 33.20 & 29.63 & 38.34 & 38.90 & 29.26 & 39.98 & 22.83 & 58.19 & 50.01 & 35.38 & 36.65\\
InternVL3.5-38B ~\cite{wang2025internvl3}  & 38B & 34.20 & 34.05 & 40.78 & 43.55 & 30.10 & 49.38 & 22.08 & 61.45 & 52.41 & 38.00 & 38.76\\
MiMo-VL-7B-RL ~\cite{coreteam2025mimovltechnicalreport} & {\ }{\ }7B & 35.14 & 38.77 & 42.25 & 50.67 & 30.15 & 49.78 & 27.22 & 66.74 & 56.25 & 39.33 & 43.80\\
MiMo-VL-7B-SFT ~\cite{coreteam2025mimovltechnicalreport}  & {\ }{\ }7B & 31.68 & 33.89 & 38.77 & 38.92 & 30.17 & 44.47 & 23.93 & 55.22 & 46.66 & 34.72 & 37.28\\
\midrule

\rowcolor[HTML]{FFE0CC}
\multicolumn{13}{c}{\itshape Specific Models} \\
\midrule
DriveLMM-o1 ~\cite{ishaq2025drivelmm} & {\ }{\ }8B & 28.89 & 29.01 & 33.21 & 35.06 & 27.54 & 37.16 & 27.00 & 52.21 & 44.72 & 33.39 & 34.33\\
RoboTron-Drive ~\cite{huang2025robotron} & {\ }{\ }8B & 30.16 & 32.18 & 27.42 & 30.23 & 18.98 & 31.31 & 13.54 & 48.61 & 43.60 & 32.28 & 30.28\\
\textbf{DSVLM (Ours)} & {\ }{\ }7B & {\bfseries 59.85} & {\bfseries 68.22} & {\bfseries 68.01} & {\bfseries 64.22} & {\bfseries 58.94} & {\bfseries 75.89} & {\bfseries 80.10} & {\bfseries 74.03} & {\bfseries 63.51} & {\bfseries 60.23} & {\bfseries 68.40} \\

\bottomrule
\end{tabular}%
}

\end{table*}

\noindent\textbf{Baselines.}
To comprehensively evaluate our benchmark, we conduct experiments on 16 VLM models, including open-source VLMs: Qwen2.5-VL~\cite{bai2025qwen2}, InternVL3.5 ~\cite{wang2025internvl3}, MiMo-VL~\cite{coreteam2025mimovltechnicalreport}, the closed-source: GPT-4o~\cite{hurst2024gpt}, Seed~\cite{guo2025seed1}, Qwen-vl~\cite{bai2023qwen}, Hunyuan-vision~\cite{sun2024hunyuan}, Glm-4v~\cite{glm2024chatglm}, and the domain-specific: DriveLMM-o1~\cite{ishaq2025drivelmm} and RoboTron-Drive~\cite{huang2025robotron}. Our findings reveal that existing models exhibit limitations in handling safety-critical tasks. To address these shortcomings, we fine-tune Qwen2.5-VL-7B using the DSBench-Dataset, resulting in a specialized assisted driving safety model, DSVLM. Table~\ref{tab:main_results} demonstrates that DSVLM achieves state-of-the-art performance across all safety-related categories.



\noindent\textbf{Evaluation Metrics.}
Since n-gram-based metrics are limited in measuring semantic correctness, we employ an LLM-based evaluation method with kimi-vl~\cite{team2025kimi} as the evaluator. This method assesses model outputs holistically in terms of semantic coherence, content completeness, and alignment with the given scene, and assigns each response a score ranging from 0 to 100.
For each dimension, the score is calculated as the average over all samples in that category. The final overall score is computed as a weighted average of these category-level scores, where the weight of each category is determined by its proportion in the dataset.

\noindent\textbf{Experimental Results.}
As shown in Table~\ref{tab:main_results}, DSVLM achieves a new state-of-the-art average score of \textbf{68.4}, outperforming the strongest commercial model, Seed-1.6 (49.52), by \textbf{18.88} points. The improvement is especially significant on challenging tasks such as cockpit understanding (\textit{Coc.}), where DSVLM reaches 80.1, substantially higher than the second-best result of 29.49. The results also suggest that instruction tuning is more critical than model scale in some cases: well-tuned open-source models, such as MiMo-VL-7B-RL, can match or even surpass larger commercial models, and within the Qwen2.5-VL series, the 32B variant outperforms the 72B one. Across tasks, \textit{Dri.} and \textit{Att.} are relatively easier, with most models achieving comparatively strong performance, whereas fine-grained and spatial reasoning tasks, particularly \textit{Coc.} and \textit{Sta.}, remain challenging. DSVLM maintains consistently strong performance in both external and in-cabin scenarios, while other models show clear degradation in the latter; this gap is especially evident on context-intensive tasks such as \textit{Coc.} and \textit{Emo.}, where DSVLM achieves 80.1 and 60.23, respectively, demonstrating its robust perception and reasoning ability in complex driving environments. More qualitative examples and visualizations are provided in the appendix.

\section{Conclusion}

In this work, we introduce \emph{driving safety understanding for assisted driving}, a new research setting that requires unified perception and reasoning over both the out-cabin traffic environment and the in-cabin driver state for safety-oriented warning. To support this task, we construct a large-scale driving safety dataset with 98K vision-language annotations and a 3K benchmark (\textbf{DSBench}) for systematic evaluation, and further develop \textbf{DSVLM}, a domain-specialized vision-language model for risk understanding and warning-oriented reasoning. Experimental results show that jointly modeling cabin-view and road-view information is essential for practical assisted driving safety, and also reveal clear limitations of both general-purpose VLMs and conventional task-specific models in diverse and long-tail safety-critical scenarios. Overall, our work establishes a unified benchmark and modeling framework for assisted driving safety, and highlights the potential of domain-specialized VLMs as effective in-vehicle safety assistants.

\begin{acks}
This work was supported by the Beijing Nova Program under Grant No. 202604841209 and the National Natural Science Foundation of China under Grant No. 62471450.
\end{acks}


\bibliographystyle{ACM-Reference-Format}
\bibliography{main}

\end{document}